%% file: iclr2027_conference.tex
\documentclass{article} % For LaTeX2e
\usepackage{iclr2027_conference,times}

\input{math_commands.tex}

\usepackage{amsmath}
\usepackage{amssymb}
\usepackage{mathtools}
\usepackage{hyperref}
\usepackage{url}
\usepackage{graphicx}
\usepackage{booktabs}    % For \toprule, \midrule, \bottomrule
\usepackage{colortbl}    % For \rowcolor
\usepackage{xcolor}      % For color definitions
\usepackage{fontawesome5}% For icons like \faIcon
\usepackage{multirow}    % If you use multirow (optional)
\newcommand{\Cow}[1][]{\includegraphics[width=10pt,trim={6cm 9cm 5cm 6cm},clip]{icons/cow.png}}
\newcommand{\Plant}[1][]{\includegraphics[width=10pt,trim={7cm 6cm 5cm 2cm},clip]{icons/plant.png}}
\newcommand{\Sheep}[1][]{\includegraphics[width=10pt,trim={6cm 7cm 5cm 6cm},clip]{icons/sheep.png}}

\newcommand{\eg}{\textit{e.g.}}
\newcommand{\ie}{\textit{i.e.}}

\usepackage{bm}

\usepackage[normalem]{ulem} 
\DeclareMathSizes{10}{8}{6}{5}
\usepackage{algorithm}
\usepackage{algorithmic}

\title{\textbf{Making MLLM Latents Matter}: Factor-Aware Predictive Learning for Cinematic Video Generation}

\author{%
  Songlin Yang$^{1,*,\dagger}$,
  Zhe Wang$^{1,*}$,
  Xiaotong Zhao$^{2}$,
  Xi Wang$^{3}$,
  Ran Zhang$^{2}$,\\
  \textbf{Shuai Li}$^{2}$,
  \textbf{Taiyi Wu}$^{2}$,
  \textbf{Songlian Li}$^{2}$,
  \textbf{Eric Liu}$^{2}$, 
  \textbf{Alan Zhao}$^{2}$, \textbf{Anyi Rao}$^{1}$ \\
  $^{1}$MMLab@HKUST, The Hong Kong University of Science and Technology \\
  $^{2}$Tencent Video, $^{3}$LIX, École Polytechnique \\
}

\iclrfinalcopy % Uncomment for camera-ready version, but NOT for submission.
\begin{document}

\maketitle

\def\thefootnote{*}\footnotetext{Equal contribution.}
\def\thefootnote{$\dagger$}\footnotetext{Project lead.}
\def\thefootnote{\arabic{footnote}}

\begin{abstract}
Cinematic video generation requires both temporal inference from limited inputs (\ie, sparse rather than frame-by-frame guidance) and factor-aware control of character, scene, action, and camera. Multimodal large language models (MLLMs) offer useful priors for both capabilities, but their representations are not automatically actionable for video generation: a Video Diffusion Transformer (DiT) may bypass them through native conditioning pathways, and using them does not ensure either factor-aware control or generation-compatible temporal prediction. We propose \textbf{Factor-Aware Predictive Learning (FAPL)}, which establishes effective MLLM latent use as a shared prerequisite and then targets the two capabilities with complementary learning signals. \textbf{First, latent-only generator grounding} trains the DiT with a frozen MLLM and connector, strengthening the DiT's use of MLLM latents for generation. \textbf{Second, for factor-aware control}, paired videos specify a target-factor change and non-target preservation, and factor-span replacement trains the DiT to realize this relationship in its output. \textbf{Third, for temporal inference}, text-only and text-plus-complete-video observations share a video target, and the trained DiT is then frozen while the MLLM and connector are optimized through it. This final stage trains limited-input representations for generation while retaining complete-observation reconstruction as a complementary training mode. Extensive experiments demonstrate our consistently better performance on three cinematic video generation tasks (\ie, sparse-to-video, factor-aware editing, and multi-shot generation) over baselines and validate FAPL's contributions to MLLM feature use, temporal control, and factor coordination. \textbf{Project Page}: \href{https://shotverse.github.io/fapl/}{https://shotverse.github.io/fapl/}
\end{abstract}

\section{Introduction}

\begin{figure}[t]
    \centering
    \includegraphics[width=\linewidth]{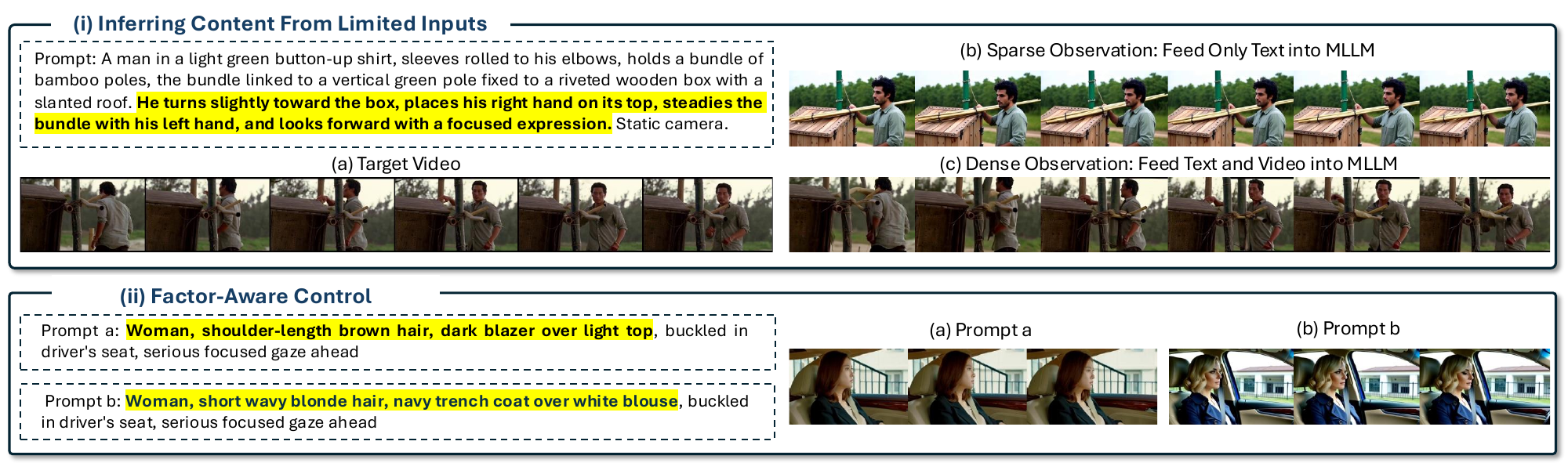}
   
    \caption{\textbf{Cinematic video generation requires inference from limited inputs and factor-aware control.} (i) The man turns toward the box, places his right hand on top, and steadies the bundle with his left hand. Text-only conditioning misses this sequence, while observing the complete video enables the same model to reproduce it. (ii) Keeping the descriptions of all other factors fixed and replacing only the character-related prompt, the non-character factors change as well. \textit{These cases illustrate failures in temporal inference and factor control, with editing providing a probe of what the generator changes and preserves.}}
    \label{fig:teaser_task}

\end{figure}

Cinematic video generation~\citep{cinemaster,shotdirector,cine_benchmark,cveu,rewardverse} helps creators explore content, performances, and camera choices before physical production. Beyond visual realism, it requires two capabilities (Fig.~\ref{fig:teaser_task}). \textbf{(i) Temporal inference from limited inputs} requires the model to construct plausible action sequences and camera motion from a script and a few visual references~\citep{captain,stage}, which only sparsely specify how a shot unfolds (\ie, they do not provide a dense, frame-by-frame specification of all visual content). \textbf{(ii) Factor-aware control} requires character, scene, action, and camera to follow their respective instructions while remaining compatible: changing a character should preserve the requested action, and changing camera motion should not replace the scene~\citep{cineweaver,cineorchestra}. Therefore, the central challenge is not only to infer missing content, but also to make those inferences selectively controllable in the generated video.

\begin{figure}[t]
    \centering
    \includegraphics[width=\linewidth]{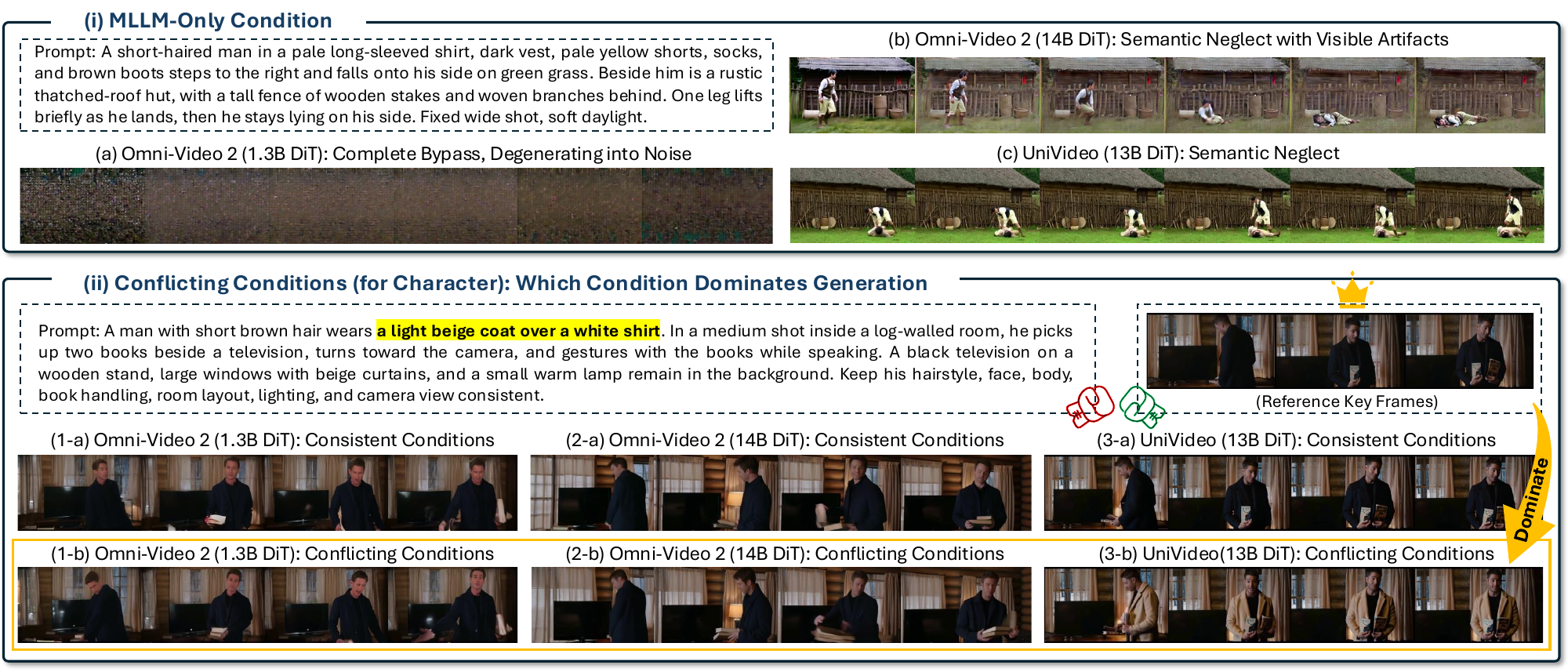}

    \caption{\textbf{MLLM latents are easily bypassed in existing MLLM--DiT models.} (i) Under MLLM-only conditioning, Omni-Video~2 (1.3B DiT) produces noise, while Omni-Video~2 (14B DiT) and UniVideo produce artifacts and depict one person's successive actions as two people acting differently. MLLM latents alone therefore fail to provide reliable generation guidance in these examples. (ii) Under conflicting conditions, VAE references dominate the output.}
    \label{fig:teaser_reason}

\end{figure}

Standard conditioning provides limited support for these requirements. Text encoders~\citep{t5,clip} offer limited visual and temporal grounding, while VAE-encoded references~\citep{vae} preserve appearance without explicitly representing action or camera intent. Multimodal large language models (MLLMs)~\citep{qwen3,qwen2} provide a promising alternative~\citep{wang2026truth}: their visual priors and semantic reasoning can help interpret scripts and references, infer plausible temporal details, and coordinate cinematic factors. Recent methods~\citep{omnivideo2,univideo} therefore condition Video Diffusion Transformers (DiTs)~\citep{dit} on projected MLLM hidden states.

However, making MLLM features available is different from making them effective for generation. Two obstacles remain: \textbf{First, MLLM feature bypass} is a shared bottleneck for both capabilities: native text-encoder and VAE-reference pathways can reduce the generation loss without substantial reliance on MLLM latents (Fig.~\ref{fig:teaser_reason}). \textbf{Second, task-structured supervision is insufficient}. For factor-aware control, ordinary video-level training does not explicitly associate an intervention on a particular MLLM span with the intended output change and non-target preservation. For temporal inference, adapting the connector and DiT around a frozen MLLM does not train the MLLM itself to produce temporally informative representations from sparse inputs. Thus, the generator must learn how to use MLLM representations, while the MLLM must also learn what representations to produce for generation.

We propose \textbf{Factor-Aware Predictive Learning (FAPL)} to strengthen the use of MLLM representations for temporal inference and factor-aware control through three training stages. \textbf{First, to strengthen the DiT's use of MLLM features}, Stage~I freezes the MLLM and connector and trains the DiT with their projected latents as the sole content conditions, using both text-only and text-plus-complete-video observations. This establishes a generation interface that subsequent stages can use for factor-aware control and predictive learning. \textbf{Second, to enable factor-aware control}, Stage~II uses paired videos that change selective cinematic factors while preserving the others. Structured descriptions locate the projected latent spans corresponding to character, scene, action, and camera. Keeping the MLLM and connector frozen, we replace the selected factor span between paired examples and train the DiT against the corresponding video target. The paired videos specify what should change and remain, while span replacement explicitly associates this visual relationship with a localized conditioning intervention, without assuming fully disentangled representations. \textbf{Finally, to learn temporal inference from sparse inputs}, Stage~III freezes the trained DiT and jointly optimizes the MLLM and connector using text-only and text-plus-complete-video observations that share a video target. The text-only mode trains prediction from limited evidence, while the complete-video mode provides complementary reconstruction supervision through the same generation interface. The fixed DiT directs learning toward generation-compatible representations without simultaneous decoder adaptation. Experiments evaluate FAPL on sparse-to-video generation, factor-aware editing, and multi-shot generation, with stage-wise ablations examining the respective contributions of strengthened MLLM feature use, factor interventions, and mixed-observation predictive learning.

Our contributions are threefold: (i) We investigate the gap between multimodal understanding and its effective use in cinematic generation, and propose \textbf{FAPL} to translate MLLM understanding into actionable guidance for inference from limited inputs and factor-aware control.  (ii) We introduce factor-aware supervision through latent interventions, turning paired edits into structured learning signals for how the generator should change and preserve cinematic content, without assuming fully disentangled representations. (iii) We formulate predictive learning through a fixed generator, linking limited-input prediction and complete-observation reconstruction via shared video targets to learn temporally informative, generation-compatible MLLM representations.

\section{Related Work}
\label{sec:related_work}

\paragraph{Cinematic Video Generation.}
Cinematic video generation~\citep{evalverse,cinesurvey,yang2026adaptive} has recently received growing attention for producing shots with specified content, performances, and camera motions from limited materials~\citep{cinemaster,holocine,shotverse}. Existing studies target the two requirements: (i) Inferring content from limited inputs. Reference-driven and multi-shot methods maintain identity and scene consistency from given references~\citep{captain, stage, phantom}. (ii) Factor-Aware Control. Camera motion, character replacement, action transfer, and scene modification are typically handled separately~\citep{motionctrl,unimoca, unic, vace, cineweaver, cineorchestra, motioncharacter,cinescene}.

\paragraph{MLLM-Conditioned Video Generation and Editing.}

Recent video diffusion models achieve high-quality synthesis and support a growing range of editing tasks~\citep{svd, hunyuanvideo, wan}. To move beyond single-task systems, recent work couples MLLMs with diffusion backbones, either through in-context conditioning~\citep{unic, fulldit, vace} or through an MLLM understanding branch connected to a generation branch~\citep{univideo, omnivideo2, omnivideo1}. Related studies improve controllability over specific factors such as camera motion~\citep{motionctrl} and subject identity~\citep{phantom}. These systems differ in architecture and task coverage, but commonly inject MLLM features alongside text-encoder and VAE reference features through a shared conditioning interface, and supervise training at the video level. Such supervision does not determine how much each condition contributes, and the generator may rely on familiar text and dense visual-reference pathways. Paired editing examples are also typically consumed as additional supervised samples, without explicitly indicating which factor should change and which content should be preserved. \textit{In contrast, FAPL strengthens the role of MLLM latents through latent-only training, and uses factor-span replacement between paired examples to supervise factor-aware changes and preservation.}

\paragraph{Predictive Representation Learning.}

Joint-embedding predictive architectures learn representations by predicting target features from partial observations rather than reconstructing pixels, avoiding unpredictable low-level details~\citep{jepa}, and extend to action-conditioned future prediction for planning and control~\citep{v-jepa,v-jepa2}. Generative models also decode predicted representations into visual outputs. However, such models usually learn task-specific spaces without high-fidelity video decoders, or require separately trained decoders, and rarely use paired editing data to supervise factor-level change and preservation. \textit{In contrast, we reuse a pretrained MLLM and Video DiT and supervise prediction through visual reconstruction with a fixed generator, preserving the decoding pathway while training the MLLM to infer video content from limited observations.}

\section{Method}
\label{sec:method}

Factor-Aware Predictive Learning (FAPL) trains MLLM representations for limited-input temporal inference and factor-aware control of character, scene, action, and camera. It structures supervision in generation space: \textbf{factor-span interventions} use paired edits to link localized latent changes to target-factor changes and non-target preservation, while \textbf{shared-target predictive learning} couples limited-input prediction with complete-observation reconstruction through a fixed generator. MLLM-only conditioning strengthens the DiT's use of these representations, supporting both mechanisms without architectural changes.

\begin{figure}[t]
    \centering
    \includegraphics[width=\linewidth]{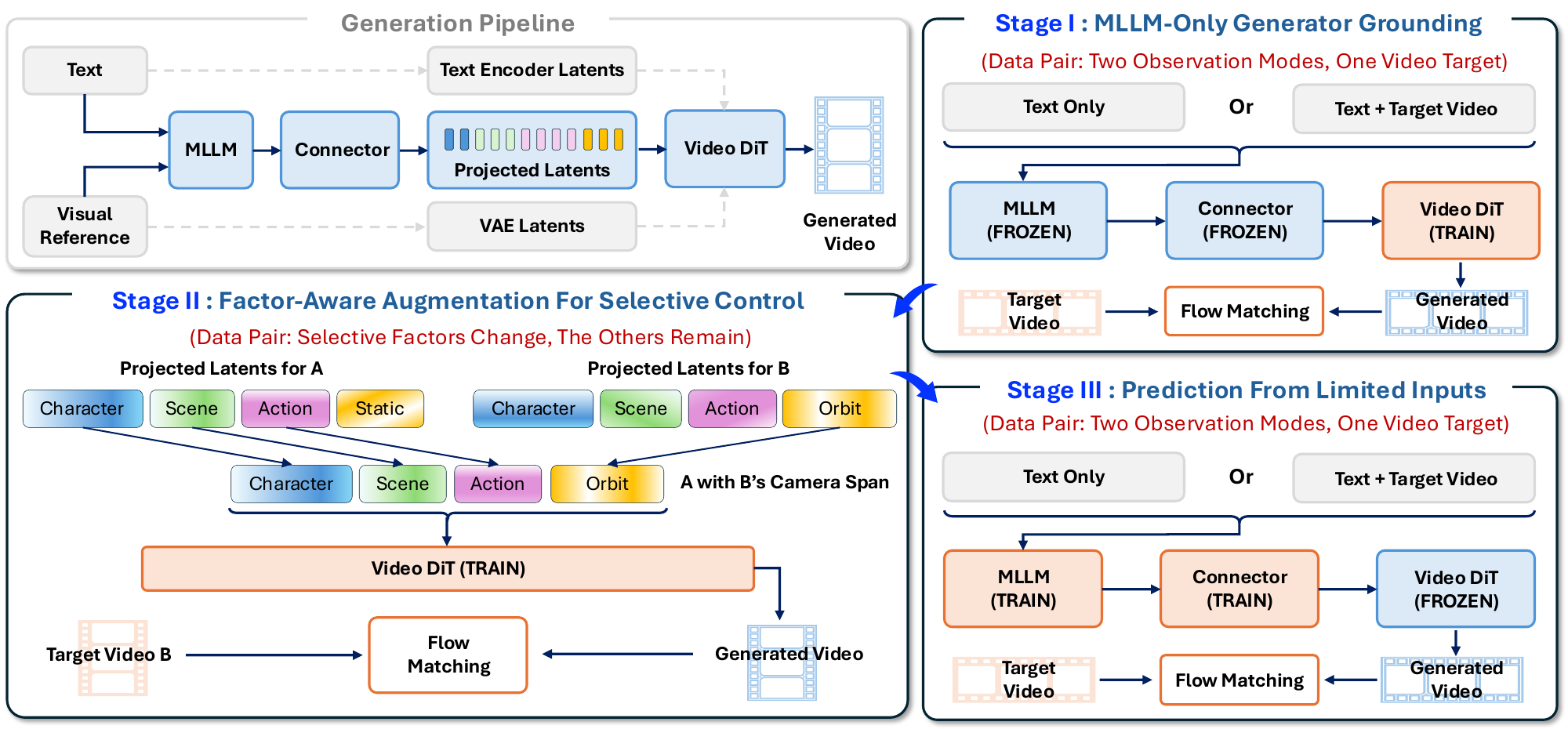}

    \caption{\textbf{Overview of Factor-Aware Predictive Learning (FAPL).} Stage I freezes the MLLM and its original connector and trains the DiT using MLLM latents. Stage II retains this configuration and swaps factor-specific output spans between paired examples to supervise factor-aware changes. Stage III freezes the trained DiT and jointly updates the MLLM and connector through video reconstruction, and text-only and text-plus-complete-video inputs share the same generation target.}
    \label{fig:framework}

\end{figure}

We first describe the framework and paired data in Sec.~\ref{sec:framework}. Training then proceeds in three stages: (i)  MLLM-Only Generator Grounding strengthens the DiT's use of MLLM features (Sec.~\ref{sec:stage1}); (ii) Factor-Aware Control Improvement trains the DiT to realize target-factor changes while preserving other content through factor-span replacement (Sec.~\ref{sec:stage2}); and (iii) Prediction From Limited Inputs optimizes the MLLM and connector through the fixed DiT using shared targets across observation modes (Sec.~\ref{sec:stage3}). All three stages use MLLM latents as the only content conditions. Fig.~\ref{fig:framework} illustrates the framework, Algorithm~\ref{alg:fapl} summarizes training, and Sec.~\ref{sec:inference} describes inference.

\subsection{Preliminary: Framework and Paired Data}
\label{sec:framework}

The generator (\ie, Video DiT) must learn both to use MLLM information and to respond appropriately in a factor-aware manner. We provide these signals through paired observations and paired edits, without introducing a new representation network.

\subsubsection{Framework: MLLM-conditioned video diffusion}

MLLM-conditioned video diffusion~\citep{omnivideo2,univideo} connects an MLLM to a pretrained Video DiT. The DiT is trained through flow matching~\citep{flow_matching} and receives three concatenated conditions: $C=[C^{\mathrm{text}};C^{\mathrm{mllm}};C^{\mathrm{ref}}]$, comprising text-encoder features, MLLM features, and VAE-encoded visual features, respectively. Here, $[\cdot;\cdot]$ denotes concatenation along the token dimension. Individual backbones may omit a pathway. For example, UniVideo~\citep{univideo} removes the original text encoders.

The MLLM condition is produced through the connector. Given a text or multimodal input $u$, the MLLM $F_{\theta}$ produces hidden states $H_{\theta}(u)=F_{\theta}(u)$, which the connector $P_{\eta}$ maps to the DiT conditioning space:
\begin{equation}
    C^{\mathrm{mllm}}
    =
    Z_{\theta,\eta}(u)
    =
    P_{\eta}\bigl(H_{\theta}(u)\bigr),
    \label{eq:mllm_condition}
\end{equation}
where $\theta$ and $\eta$ are the MLLM and connector parameters. We retain the original hidden-state extraction and connector. Our factor-level interventions operate on the projected sequence $Z_{\theta,\eta}(u)$, after the connector and before the DiT.

For a target video $x$ with VAE latent $z_0$, flow matching constructs $z_t=\alpha_tz_0+\sigma_t\epsilon$, where $\epsilon\sim\mathcal{N}(0,I)$ is Gaussian noise, $t$ is the flow time, and $\alpha_t,\sigma_t$ specify the interpolation schedule. The DiT $G_{\phi}$, with parameters $\phi$, predicts the target velocity $v^*=\partial z_t/\partial t$:
\begin{equation}
    \mathcal{L}_{\mathrm{FM}}(x\mid C;\phi)
    =
    \mathbb{E}_{t,\epsilon}
    \left[
        \left\|
            G_{\phi}(z_t,t\mid C)-v^*
        \right\|_2^2
    \right].
    \label{eq:flow_matching}
\end{equation}
We follow the pretrained backbone's flow schedule and sampling procedure. This objective supervises generation under the combined conditions, but does not determine their individual contributions. The DiT may rely primarily on text and VAE features, so accurate reconstruction alone does not establish effective use of MLLM latents.

\begin{algorithm}[t]
    \caption{Three-Stage Training of Factor-Aware Predictive Learning (FAPL)}
    \label{alg:fapl}
    \footnotesize
    \begin{algorithmic}[1]
        \REQUIRE Pretrained MLLM $F_{\theta_0}$, connector $P_{\eta_0}$, and DiT $G_{\phi_0}$; observation data $\mathcal{D}_{\mathrm{obs}}$ and editing pairs $\mathcal{D}_{\mathrm{edit}}$.
        \STATE Initialize $\theta\leftarrow\theta_0$, $\eta\leftarrow\eta_0$, and $\phi\leftarrow\phi_0$; Use only MLLM latents as content conditions in all stages.
        \STATE \textbf{Stage I: Enhance MLLM Latent Use (MLLM-Only Generator Grounding).}
        \FOR{each training step}
            \STATE Sample $(x,p)\sim\mathcal{D}_{\mathrm{obs}}$ and observation mode $m\sim\rho$; Compute $Z_{\theta_0,\eta_0}(u^m)$.
            \STATE Update $\phi$ using $\mathcal{L}_{\mathrm{FM}}(x\mid Z_{\theta_0,\eta_0}(u^m);\phi)$; keep $\theta_0,\eta_0$ fixed.
        \ENDFOR
        \STATE \textbf{Stage II: Improve Factor-Aware Control.}
        \FOR{each training step}
            \STATE Sample $(x^a,p^a,x^b,p^b,k)$ from $\mathcal{D}_{\mathrm{edit}}$; Compute $Z^a,Z^b$ and locate their factor-$k$ spans.
            \STATE Replace the designated spans as in Eq.~\ref{eq:factor_swap}; Update $\phi$ using $\mathcal{L}_{\mathrm{edit}}$; keep $\theta_0,\eta_0$ fixed.
        \ENDFOR
        \STATE Save $\phi^{*}\leftarrow\phi$ and freeze the DiT.
        \STATE \textbf{Stage III: Learn Prediction From Limited Inputs.}
        \FOR{each training step}
            \STATE Sample $(x,p)$ from $\mathcal{D}_{\mathrm{obs}}$ and observation mode $m\sim\pi$.
            \STATE Construct $u^{\mathrm{t}}=p$ if $m=\mathrm{t}$, and $u^{\mathrm{tv}}=(p,x)$ otherwise.
            \STATE Compute $Z_{\theta,\eta}(u^m)$ and update $\theta,\eta$ using
            $\mathcal{L}_{\mathrm{FM}}
            (x\mid Z_{\theta,\eta}(u^m);\phi^{*})$;
            keep $\phi^{*}$ fixed.
        \ENDFOR
    \end{algorithmic}
\end{algorithm}

\subsubsection{Paired Data}

\paragraph{Two Observation Modes, One Video Target.}
For each training video $x$ and its description $p$, we construct two inputs: $u^{\mathrm{t}}=p$ and $u^{\mathrm{tv}}=(p,x)$. Both are supervised with the same target video $x$. The text-only input requires inference from a description, while the text-plus-video input provides direct visual evidence. We denote these examples by $\mathcal{D}_{\mathrm{obs}}$. The complete target video is never supplied to the text-only branch.

\paragraph{Selective Factors Change, the Others Remain.}
We construct the editing set $\mathcal{D}_{\mathrm{edit}}$ through factor-specific selection and generation of source--target video pairs, covering changes to character, scene, action, and camera motion. Each tuple $(x^a,p^a,x^b,p^b,k)$ contains two videos $x^a,x^b$ and their descriptions $p^a,p^b$, with only the selected factor $k\in\mathcal{F}=\{\mathrm{character},\mathrm{scene},\mathrm{action},\mathrm{camera}\}$ modified. Both descriptions are structured prompts: we rewrite each caption into four ordered fields, character, scene, action, and camera, and obtain $p^b$ by updating only the edited factor's field while retaining the others. Field labels and boundaries locate the corresponding token spans in the MLLM hidden states and projected latents, giving each factor an explicit address for intervention. Span lengths follow the text and need not match across examples, and we do not assume the contextualized spans are fully disentangled.

\subsection{Stage I: MLLM-Only Generator Grounding}
\label{sec:stage1}

MLLM representations can influence generation only if the DiT can reliably decode and use them. In the original architecture, native text-encoder and VAE-reference pathways allow the DiT to reduce the generation loss without sufficiently relying on MLLM latents. Stage~I therefore isolates the MLLM pathway and trains the DiT under both limited and complete observations. We freeze the pretrained MLLM and connector and update only the DiT.

\paragraph{Mixed-Observation Latent-Only Grounding.} For each video $x$ and description $p$, we use the text-only input $u^{\mathrm{t}}=p$ and the text-plus-complete-video input $u^{\mathrm{tv}}=(p,x)$. The frozen MLLM and connector produce $Z_{\theta_0,\eta_0}(u^m)$ for observation mode $m\in\{\mathrm{t},\mathrm{tv}\}$. In both modes, the DiT receives MLLM latents as its only content condition; the separate text-encoder and VAE-reference conditions are disabled. Dense observations ground the DiT in the visual and temporal information carried by MLLM latents, while text-only observations maintain compatibility with limited-input generation and the text-based interventions in Stage~II.

\paragraph{Mixed-Observation Objective.}
Stage~I minimizes
\begin{equation}
    \mathcal{L}_{\mathrm{ground}}(\phi)
    =
    \mathbb{E}_{(x,p)\sim\mathcal{D}_{\mathrm{obs}}}
    \left[
        \sum_{m\in\{\mathrm{t},\mathrm{tv}\}}
        \rho_m\,
        \mathcal{L}_{\mathrm{FM}}
        \left(
            x
            \mid
            Z_{\theta_0,\eta_0}(u^m);
            \phi
        \right)
    \right],
    \label{eq:mixed_grounding_loss}
\end{equation}
where $\rho_{\mathrm{t}}+\rho_{\mathrm{tv}}=1$. We optimize $\phi$ while keeping $\theta_0$ and $\eta_0$ fixed. Because alternative content pathways are unavailable, the DiT must learn to use MLLM latents across both observation modes. Stage~I adapts the generator to fixed MLLM representations, while Stage~III later reverses this direction and adapts the MLLM pathway through a fixed generator.

\subsection{Stage II: Factor-Aware Control Improvement}
\label{sec:stage2}

Using MLLM features is not sufficient for factor-aware control. The DiT must also learn which output changes correspond to a particular change in its conditions. Stage II introduces factor-level latent augmentation with explicit video targets. The MLLM and connector remain frozen, and training uses text-only inputs with MLLM-latent-only conditioning.

\paragraph{Factor-Span Replacement.}
For a pair in $\mathcal{D}_{\mathrm{edit}}$, let
$Z^a=Z_{\theta_0,\eta_0}(p^a)$ and
$Z^b=Z_{\theta_0,\eta_0}(p^b)$
be the projected latents of the two descriptions.
Define $\mathcal{R}_k(A,B)$ as replacing the factor-$k$ span in $A$ with the corresponding span from $B$, leaving the remaining tokens unchanged.
The augmented conditions are
\begin{equation}
    Z_k^{a\leftarrow b}
    =
    \mathcal{R}_k(Z^a,Z^b),
    \qquad
    Z_k^{b\leftarrow a}
    =
    \mathcal{R}_k(Z^b,Z^a).
    \label{eq:factor_swap}
\end{equation}
Replacement is applied after the connector and before the DiT, so no additional projection follows.
Complete spans are removed and inserted, so replacement does not require equal token counts; sequence lengths and masks are updated accordingly.
Since the pair changes only factor $k$, the condition $Z_k^{a\leftarrow b}$ is supervised with $x^b$, and the reverse condition is supervised with $x^a$.

\paragraph{Editing Objective.}
We update the DiT using both replacement directions:
\begin{equation}
    \mathcal{L}_{\mathrm{edit}}(\phi)
    =
    \mathbb{E}_{(x^a,p^a,x^b,p^b,k)\sim\mathcal{D}_{\mathrm{edit}}}
    \left[
        \mathcal{L}_{\mathrm{FM}}
        \bigl(x^b\mid Z_k^{a\leftarrow b};\phi\bigr)
        +
        \mathcal{L}_{\mathrm{FM}}
        \bigl(x^a\mid Z_k^{b\leftarrow a};\phi\bigr)
    \right].
    \label{eq:editing_loss}
\end{equation}
The target video supervises the effect of the replaced span and preservation of the remaining factors. Thus, factor awareness is learned through the generator's response to controlled replacements, rather than through a feature-separation objective. We denote the DiT parameters after this stage by $\phi^{*}$.

\subsection{Stage III: Prediction From Limited Inputs}
\label{sec:stage3}

The first two stages establish a generator that uses MLLM features and responds to factor-aware interventions. Stage~III uses this fixed generator to optimize the MLLM pathway itself. The goal is to produce generation-compatible representations from limited observations, while complete-video observations provide a richer training mode for the same video target.

\paragraph{Shared Targets Across Observation Modes.}
We freeze the DiT at $\phi^{*}$ and jointly fine-tune the MLLM and connector, initialized from $\theta_0$ and $\eta_0$. For each video $x$ and description $p$, we use the text-only input
$u^{\mathrm{t}}=p$
and the text-plus-complete-video input
$u^{\mathrm{tv}}=(p,x)$.
Both observation modes use MLLM-latent-only conditioning and are supervised with the same target video $x$.

We define the mixed-observation objective as
\begin{equation}
    \mathcal{L}_{\mathrm{obs}}(\theta,\eta)
    =
    \mathbb{E}_{(x,p)\sim\mathcal{D}_{\mathrm{obs}}}
    \left[
        \sum_{m\in\{\mathrm{t},\mathrm{tv}\}}
        \pi_m\,
        \mathcal{L}_{\mathrm{FM}}
        \left(
            x
            \mid
            Z_{\theta,\eta}(u^m);
            \phi^{*}
        \right)
    \right],
    \label{eq:observation_loss}
\end{equation}
where $\pi_m$ denotes the sampling weight of observation mode $m$, with
$\pi_{\mathrm{t}}+\pi_{\mathrm{tv}}=1$.
We minimize Eq.~\ref{eq:observation_loss} over $\theta$ and $\eta$, while keeping $\phi^{*}$ fixed.

\paragraph{Prediction Through a Fixed Generator.}
Although the DiT is frozen, gradients still pass through it to the connector and MLLM. Under text-only observation, the MLLM pathway must produce representations that let the fixed generator reconstruct the target video's semantic, motion, and temporal content from limited evidence. Under complete-video observation, the MLLM has direct visual evidence but must express it through the same connector and generation interface. Training both modes toward a shared video target therefore adapts the MLLM pathway without changing the DiT.

\subsection{Inference and Cinematic Control}
\label{sec:inference}

At inference, FAPL supports three cinematic applications within one architecture. \textbf{(i) Generation from Scripts and References.} Given a script and optional character or scene references, the MLLM produces latents that condition the DiT through the trained connector and original interface, followed by flow-based sampling and VAE decoding. No complete target video is needed. The DiT can use MLLM latents alone or together with the original text and VAE reference conditions as separate inference configurations. \textbf{(ii) Factor-Aware Control.} We modify only the target factor's field in the structured prompt and re-encode the edited prompt through the MLLM and connector. We then replace the corresponding span in the original projected latents with the edited field's span, keeping all other spans unchanged, and condition the DiT on the resulting sequence. We evaluate both the requested change and non-target preservation. \textbf{(iii) Multi-Shot Generation.} Shots are generated independently from their descriptions and available references. Shared character and scene references promote consistency, while action and camera instructions vary by shot.

\section{Experiments}
\label{sec:experiments}

\subsection{Implementation}
We build FAPL on UniVideo~\citep{univideo} with a 13B DiT and compare with the original UniVideo and Omni-Video~2~\citep{omnivideo2} models (1.3B, 14B DiTs). Stages~I and III sample text-only and text-plus-complete-video observations at a 1:1 ratio, while Stage~II uses text-only inputs with factor-span replacement. We train Stages~I--III for 10K, 15K, and 10K steps, respectively, using AdamW with a learning rate of $1\times10^{-5}$, zero weight decay, and a maximum gradient norm of 1.0. Training uses the bfloat16 (BF16) format and gradient checkpointing. Inference uses 50 sampling steps and a classifier-free guidance (CFG) scale of 5.0. All three stages share 100K quality-filtered source--target video pairs (25K per factor), constructed through factor-specific selection and generation using CoinVE-200K~\citep{coinve} as the primary source and OmniVBench's released training data~\citep{OmniVBench} for supplementary videos and references.

\subsection{Evaluation Data, Tasks, and Metrics}

\paragraph{Test Data.} CineFactor-10K comprises 10K single-shot test cases constructed from held-out source groups in CoinVE-200K~\citep{coinve} and the training dataset released with OmniVBench~\citep{OmniVBench}, with no source-video overlap with our training data.

\paragraph{Tasks.} (i) Sparse-to-Video Generation. We evaluate on 5K cases using text-only prompts. (ii) Factor-Aware Editing. We evaluate on 5K cases, with 1.25K cases per factor. Each case modifies only the target factor's field in the structured prompt while retaining the other three fields unchanged. All methods receive the same modified prompt and permitted visual inputs through their standard inference interfaces. (iii) Multi-Shot Generation. We evaluate separately on the 100 sequences in the official ShotVerse-Bench~\citep{shotverse} test split. Each sequence comprises two to four shots in their original order, with shared character and scene references and shot-specific action and camera instructions.

\paragraph{Metrics.} (i) Temporal generation from limited inputs. Within shots, we measure action sequencing with Motion Order~\citep{vbench2} and camera-trajectory fidelity with Translation Error and Rotation Error (radians)~\citep{shotverse}, using frozen $\pi^3$~\citep{pi3} for trajectory recovery. To complement these action/camera measures, multi-shot evaluation assesses character and scene consistency as shot-specific instructions vary, using Subject Consistency, Scene Consistency~\citep{multishotmaster}, and FaceSim-Arc for facial identity~\citep{yuan2025identity}. (ii) Factor-aware control. We evaluate single-factor edits using FiVE's $\cap$-Acc~\citep{five}, requiring both yes/no and multiple-choice checks to pass.

\subsection{Comparison and Analysis}
\label{sec:comparison}

\begin{figure}[t]
    \centering
    \includegraphics[width=\linewidth]{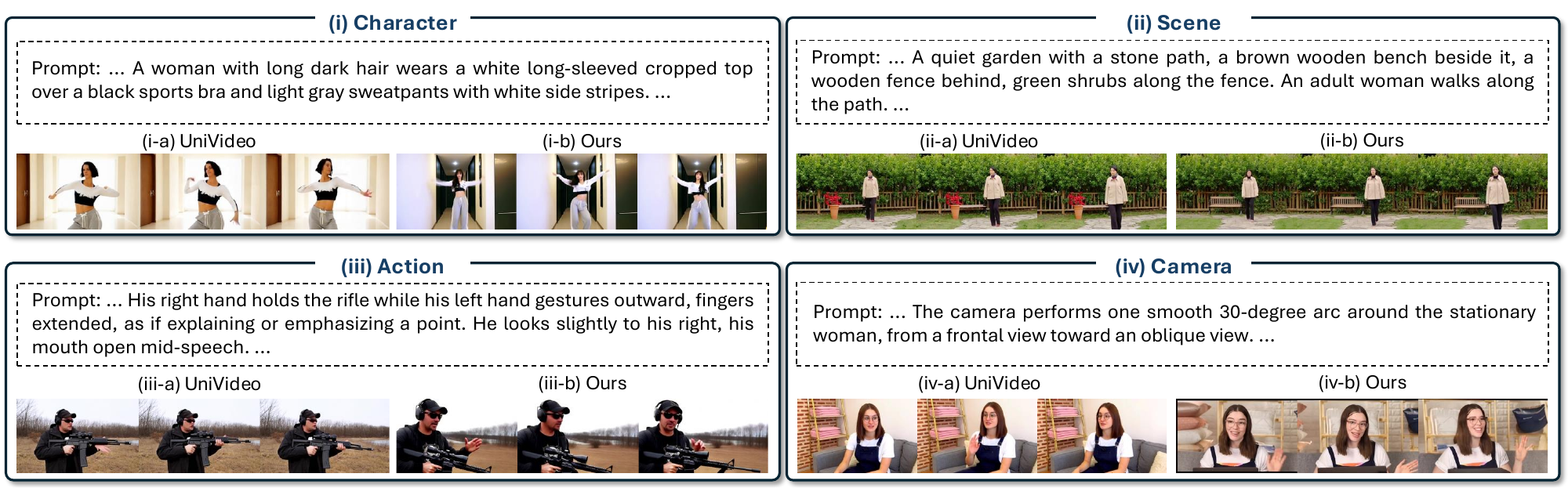}

    \caption{
    \textbf{Under limited inputs, baseline models fail to predict temporal visual details.} With only text and no visual references, we compare UniVideo with FAPL (Ours), highlighting representative baseline limitations across four cinematic factors. \textbf{(i) Character.} Long hair is generated as short hair, failing appearance details specified only by text. \textbf{(ii) Scene.} The background is not preserved and incorrect content appears. \textbf{(iii) Action.} The requested sequence, \eg, gesturing outward with the left hand and extended fingers, is missing. \textbf{(iv) Camera.} UniVideo ignores the instruction and keeps the camera static, while Ours orbits around the character. \textit{FAPL better realizes prompt-specified appearance, scene, and motion from text alone.}}
    \label{fig:sparse-to-video}

\end{figure}

\begin{figure}[t]
    \centering
    \includegraphics[width=\linewidth]{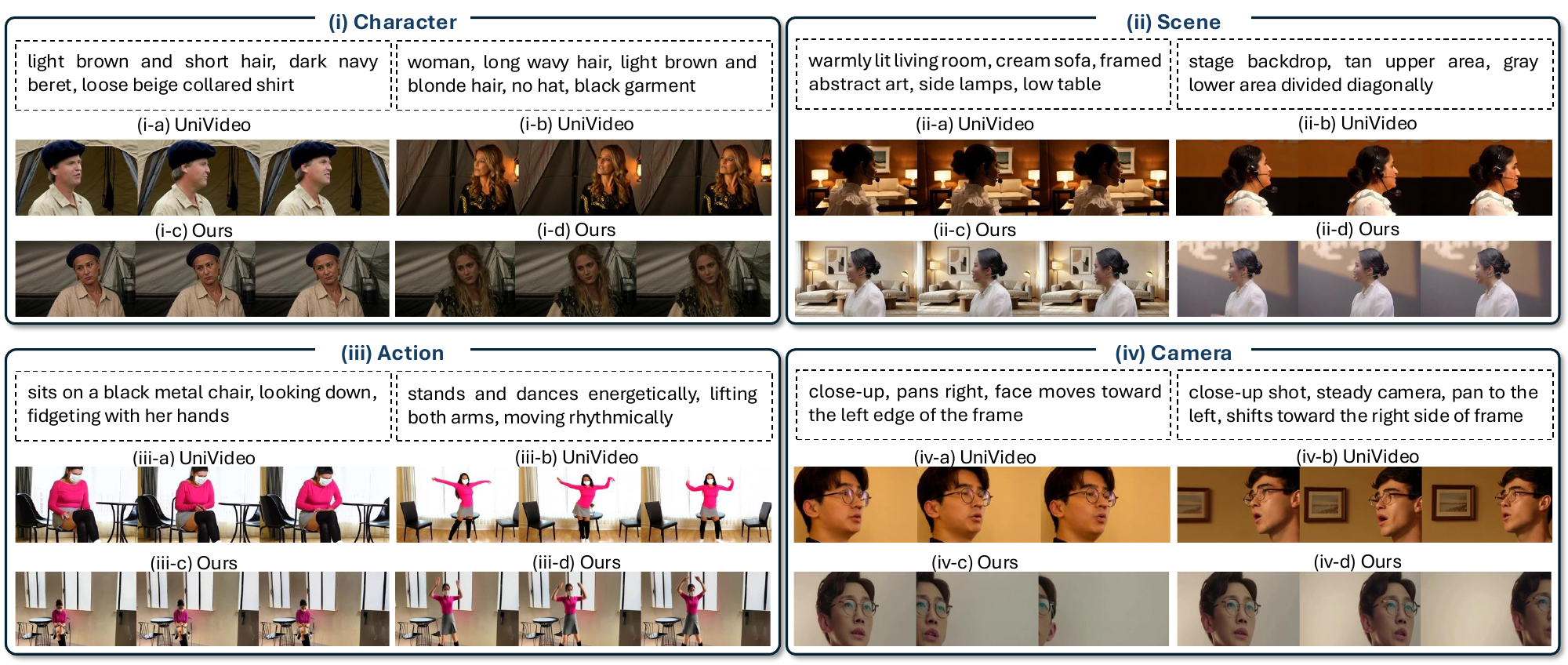}

    \caption{
    \textbf{FAPL improves factor-aware control.}
    For each factor, we modify only the corresponding description in the structured prompt and generate twice, comparing UniVideo with FAPL (Ours). UniVideo does not confine the change to the requested factor and alters unrelated content, whereas Ours modifies the selected factor while preserving the others. \textit{This indicates that factor-level latent augmentation enables selective control.}}
    \label{fig:factor-aware}

\end{figure}

\begin{figure}[t]
    \centering
    \begin{minipage}[t]{0.495\textwidth}
        \centering
        \includegraphics[width=\linewidth]{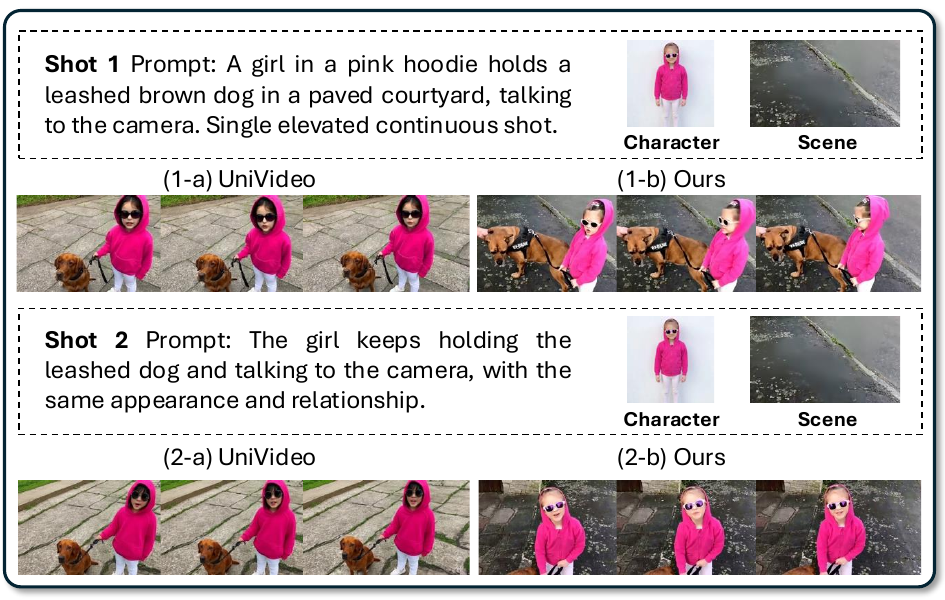}
        \vspace{-0.8cm}
        \caption{
    \textbf{FAPL supports consistent multi-shot generation without cross-shot memory.}
}
        \label{fig:multishot}
    \end{minipage}
    \hfill
    \begin{minipage}[t]{0.495\textwidth}
        \centering
        \includegraphics[width=\linewidth]{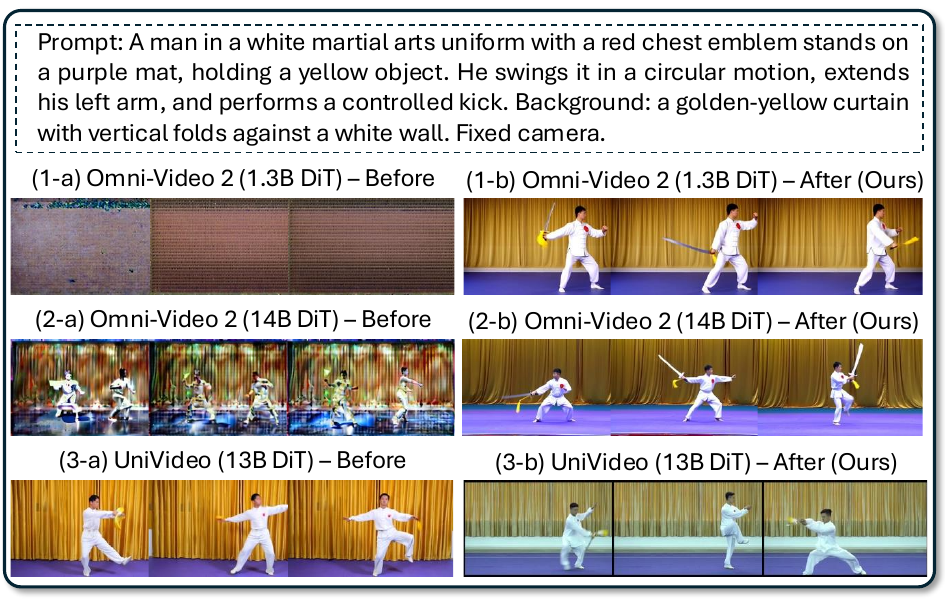}
        \vspace{-0.8cm}
        \caption{\textbf{MLLM latent enhancement generalizes across MLLM--DiT models.}}
        \label{fig:mllm-only-ablation}
    \end{minipage}

\end{figure}

Tab.~\ref{tab:main_comparison} compares FAPL with Omni-Video~2 and UniVideo on sparse-to-video generation, factor-aware editing, and multi-shot generation. FAPL improves action ordering and camera-trajectory fidelity from text alone, achieves higher $\cap$-Acc for requested factor changes, and obtains stronger Subject and Scene Consistency across shots, indicating that shot-specific instructions remain compatible with shared character and scene content. For multi-shot generation, FAPL achieves a FaceSim-Arc score of 0.523, compared with 0.412 for Omni-Video~2 (1.3B), 0.468 for Omni-Video~2 (14B), and 0.451 for UniVideo (13B). This improvement complements the Subject and Scene Consistency results, indicating stronger facial identity consistency as action and camera instructions vary across shots. In the examples shown in Figs.~\ref{fig:sparse-to-video}, \ref{fig:factor-aware}, and~\ref{fig:multishot}, UniVideo captures the broad content but misses specified details such as hair length, hand gestures, or the requested orbit, and some edits also alter non-target factors. FAPL follows these specifications more closely while preserving unrelated factors, and maintains the shared setting and character appearance across shots. These comparisons suggest that FAPL improves instruction following and cross-shot consistency on the evaluated tasks.

\begin{table}[t]
    \centering
\caption{\textbf{Quantitative comparison across three evaluation tasks.} }
    \label{tab:main_comparison}
    \small
    \setlength{\tabcolsep}{4pt}
    \renewcommand{\arraystretch}{1.1}
    \resizebox{\linewidth}{!}{%
    \begin{tabular}{lccccccc}
        \toprule
        & \multicolumn{3}{c}{Sparse-to-Video}
        & \multicolumn{1}{c}{Factor-Aware Editing}
        & \multicolumn{3}{c}{Multi-Shot Consistency} \\
        \cmidrule(lr){2-4}
        \cmidrule(lr){5-5}
        \cmidrule(lr){6-8}
        Method
        & Motion Order (\%)$\uparrow$
        & Trans. Error$\downarrow$
        & Rot. Error (rad)$\downarrow$
        & $\cap$-Acc (\%)$\uparrow$
        & Subject$\uparrow$
        & Scene$\uparrow$
        & FaceSim-Arc$\uparrow$ \\
        \midrule
        Omni-Video~2 (1.3B)
        & 33 & 0.1084 & 1.87 & 28 & 0.435 & 0.373 & 0.412 \\
        Omni-Video~2 (14B)
        & 43 & 0.0817 & 1.49 & 37 & 0.476 & 0.423 & 0.468 \\
        UniVideo (13B)
        & 48 & 0.0763 & 1.54 & 42 & 0.462 & 0.427 & 0.451 \\
        \midrule
        \textbf{FAPL (Ours)}
        & \textbf{61} & \textbf{0.0586} & \textbf{1.02}
        & \textbf{58} & \textbf{0.516} & \textbf{0.471}
        & \textbf{0.523} \\
        \bottomrule
    \end{tabular}%
    }

\end{table}

\subsection{Ablation Study}
\label{sec:ablation}

Tab.~\ref{tab:ablation} evaluates the three training stages on UniVideo (13B). Stage~I improves action ordering and reduces both camera errors relative to the pretrained model. Fig.~\ref{fig:mllm-only-ablation} further illustrates improved generation under MLLM-only conditioning after Stage~I on Omni-Video~2 (1.3B and 14B) and UniVideo (13B). For Stage~II, training with factor swapping achieves 57\% $\cap$-Acc, compared with 48\% for training without factor swapping under the same paired data and optimization budget. Motion Order and camera errors do not improve over this control, suggesting that the benefit is concentrated in factor-aware editing. For Stage~III, mixed-observation training improves Motion Order from 57\% to 61\% and reduces translation and rotation errors from 0.0629 to 0.0586 and from 1.26 to 1.02 radians, respectively, compared with text-only training. Editing accuracy remains comparable to the Stage~II checkpoint, changing from 57\% to 58\%. These results suggest that complete-video observations during training help action ordering and camera control under prompt-only inference while retaining factor-aware editing performance.

\begin{table}[H]
    \centering
    \small
    \setlength{\tabcolsep}{2pt}
    \renewcommand{\arraystretch}{1.15}
    \caption{\textbf{Ablation of FAPL training stages on UniVideo (13B).}}
    \label{tab:ablation}
    \resizebox{\linewidth}{!}{%
    \begin{tabular}{clcccc}
        \toprule
        Exp Index
        & Training Configuration
        & Motion Order (\%)$\uparrow$
        & Trans. Error$\downarrow$
        & Rot. Error (rad)$\downarrow$
        & $\cap$-Acc (\%)$\uparrow$ \\
        \midrule
        0 &Pretrained UniVideo
        & 48 & 0.0763 & 1.54 & 42 \\
        1& Stage I only
        & 54 & 0.0708 & 1.38 & 44 \\
        \midrule
        2&Stage II: training without factor swapping (based on Exp.~1)
        & 54 & 0.0654 & 1.30 & 48 \\
        3& Stage II: training with factor swapping (based on Exp.~1)
        & 53 & 0.0671 & 1.32 & 57 \\
        \midrule
        4&Stage III: text-only training (based on Exp.~3)
        & 57 & 0.0629 & 1.26 & 56 \\
        5&Stage III: (Full FAPL) mixed-observation training (based on Exp.~3)
        & \textbf{61} & \textbf{0.0586}
        & \textbf{1.02} & \textbf{58} \\
        \bottomrule
    \end{tabular}%
    }

\end{table}

\section{Conclusion}
This work studies how MLLM representations can support temporal inference and factor-aware control in cinematic video generation. FAPL makes these representations actionable by grounding the DiT in MLLM latents, supervising factor-specific changes and preservation, and learning prediction from limited inputs. FAPL demonstrates that targeting the right bottleneck delivers substantial gains without added architectural complexity. Our findings further offer insights into how multimodal reasoning and generation can reinforce each other: reasoning guides controllable synthesis, while generation provides grounded supervision for learning to infer missing content from limited observations.

\paragraph{Limitations and Future Work.} Although FAPL addresses a fundamental weakness of existing MLLM-conditioned DiTs by making MLLM latents more actionable, its performance remains bounded by the underlying models. Current MLLMs struggle to interpret 3D blockouts or reason about complex camera motion for DiT control, while the pretrained DiT limits visual fidelity and complex factor realization, such as realistic acting. Future work may incorporate stronger MLLMs and more capable video generators. 

\bibliography{iclr2027_conference}
\bibliographystyle{iclr2027_conference}

\end{document}

%% file: math_commands.tex
\usepackage{amsmath,amsfonts,bm}

\def\eqref#1{equation~\ref{#1}}
\def\1{\bm{1}}

\DeclareMathAlphabet{\mathsfit}{\encodingdefault}{\sfdefault}{m}{sl}
\SetMathAlphabet{\mathsfit}{bold}{\encodingdefault}{\sfdefault}{bx}{n}